\documentclass[conference]{IEEEtran}

\IEEEoverridecommandlockouts
\usepackage{cite}
\usepackage{amsmath,amssymb,amsfonts, amsthm}
\usepackage{algorithmic, algorithm, }
\usepackage{threeparttable}
\usepackage{graphicx}
\usepackage{textcomp}
\usepackage{xcolor}
\usepackage{subfigure}
\usepackage{booktabs}
\usepackage{array}
\usepackage{multirow}
\usepackage{setspace}
\usepackage{caption}

\def\BibTeX{{\rm B\kern-.05em{\sc i\kern-.025em b}\kern-.08em
    T\kern-.1667em\lower.7ex\hbox{E}\kern-.125emX}}
\begin{document}

\title{AWD3: Dynamic Reduction of the Estimation Bias}

\author{\IEEEauthorblockN{Dogan C. Cicek\textsuperscript{*}, Enes Duran\textsuperscript{*}, Baturay Saglam, Kagan Kaya, Furkan Mutlu, Suleyman S. Kozat$\dagger$} 
\IEEEauthorblockA{\textit{Electrical and Electronics Engineering Department, Bilkent University, Ankara, Turkey}} \IEEEauthorblockA{ \{cicek; enesd; baturay; burak.mutlu; kozat\}@ee.bilkent.edu.tr; kagan.kaya@ug.bilkent.edu.tr}
\IEEEauthorblockA{\textsuperscript{*}Equal contribution, $\dagger$IEEE Senior Member}
}

\maketitle 
\begin{abstract}
    Value-based deep Reinforcement Learning (RL) algorithms suffer from the estimation bias primarily caused by function approximation and temporal difference (TD) learning. This problem induces faulty state-action value estimates and therefore harms the performance and robustness of the learning algorithms. Although several techniques were proposed to tackle, learning algorithms still suffer from this bias. Here, we introduce a technique that eliminates the estimation bias in off-policy continuous control algorithms using the experience replay mechanism. We adaptively learn the weighting hyper-parameter beta in the Weighted Twin Delayed Deep Deterministic Policy Gradient algorithm. Our method is named Adaptive-WD3 (AWD3). We show through continuous control environments of OpenAI gym that our algorithm matches or outperforms the state-of-the-art off-policy policy gradient learning algorithms.

\end{abstract}


\section{Introduction}
Reinforcement learning (RL) studies how an agent interacts with its environment to learn a good policy via optimizing the cumulative sum of delayed rewards. This field gained considerable attention recently due to exciting developments such as controlling continuous systems \cite{mnih2015human} and achieving superhuman level performance in Atari games \cite{vanhasselt2015deep}. Despite these promising developments, deep RL agents suffer from some issues precluding agents from performing a broad range of tasks \cite{thrun1993issues}. One of the well-known problems is the estimation bias in value-based deep RL algorithms \cite{fujimoto2018addressing, DBLP:journals/corr/abs-1812-02648}.

Estimation bias (overestimation and underestimation) is the chronic case of predicting faulty values diverging from the true values. Using neural networks as function approximators leads to an unavoidable noise due to the imprecision of the estimator. Temporal difference learning -using subsequent value estimation to update current estimate- further amplifies this bias\cite{s-lpmtd-88}. Recent works show that Q-learning, a TD learning method, is susceptible to overestimation originating from the maximization operation of noisy value estimates.   \cite{DBLP:journals/corr/abs-1812-02648, thrun1993issues}. 

The estimation bias in discrete action spaces has been widely studied \cite{lan2020maxmin, anschel2017averageddqn}. Hasselt et al. reveals that using a single value estimator causes overestimation. They propose the Double DQN algorithm by introducing an additional function estimator, named target network \cite{vanhasselt2015deep} that proved to be more effective than the vanilla version. Similarly, the DDPG algorithm uses a target network to eliminate overestimation for continuous control tasks. Despite its effectiveness in some tasks, the Q-value estimator network (critic) is susceptible to overestimation, especially in large state spaces \cite{mnih2015human}.

On the other hand, Fujimoto et al. introduced the TD3 algorithm through an additional critic network and taking the minimum of the network pair to estimate action-value function \cite{fujimoto2018addressing}. Although minimization presents an effective way to handle the overestimation, it introduces underestimation in some cases. The agent's pessimism about state values often harms its performance. In response to this, He et al. proposed a weighted average of the two terms, minimum and average of the two approximators. This proposition (WD3 algorithm) improved the precision of action-value estimators and contributed to the learning process \cite{DBLP:conf/ictai/HeH20}. Unfortunately, it introduces a weighting hyper-parameter $\beta$ which is challenging to tune. Moreover, the proposed range of the weighting hyper-parameter (0,1) is insufficient, such that in some cases even taking the minimum or the average of the two value approximators leads to the estimation bias. Additionally, having a fixed weighting value does not cover the non-stationary environments because it implicitly assumes that this weighting ratio is valid throughout the whole learning process. Hence, a further study of the estimation bias in continuous action spaces is needed.

Here, we focus on the problem of the estimation bias in the continuous action spaces. We present a technique to update the weighting hyper-parameter $\beta$ in the WD3 algorithm. Our main motivation is to dynamically combine two opposite sides to balance the estimation bias. In this way, the need for tuning the weighting hyper-parameter is eliminated.  We named our algorithm Adaptive Weighted Delayed Deep Deterministic Policy Gradient  (AWD3). Our major contributions are:

\begin{itemize}
\item We empirically demonstrate the estimation bias in the former algorithms. We show the necessity to further expand the range of the weighting hyper-parameter $\beta$. 
\item We introduce a mechanism to update $\beta$. Through simulations on OpenAI gym environments, we show that our approach performs better than other state-of-the-art policy gradient algorithms. In these simulations, our approach better estimates the state-action values than the predecessor algorithms. 
\end{itemize}

\section{Background}

We consider the standard RL paradigm where an agent interacts with an environment in discrete time-steps to learn reward-optimal behavior \cite{Sutton1998}. The standard RL paradigm is formalized as a Markov Decision Process (MDP). Each discrete time-step $t$ with a given state $s \! \in \! S$, the agent selects an action $a \! \in \! A(s_t)$ according to its policy $\pi$, receives a reward signal $r$ and the new state $s^{\prime}$ from the environment. The return is the discounted sum of rewards $ R_{t}\!\!=\!\!\sum_{i=t}^{T} \gamma^{i-t} r\left(s_{i}, a_{i}\right)
$ where $\gamma$ is the discount factor denoting the importance given to the future rewards. Next subsections introduce related policy gradient RL algorithms.

\subsection{Deep Deterministic Policy Gradient (DDPG)}
 
DDPG algorithm combines the Deterministic Policy Gradient algorithm with function approximation \cite{lillicrap2019continuous}. DDPG agents learn a Q function with a valid policy sampled in continuous space. The algorithm contains four neural networks named actor and critic along with their target networks. The critic network is used to learn the Q function. The actor network is responsible for outputting a deterministic action given the state ($ \pi\!\!:\!\! S\!\!\rightarrow\!\! A $).  The actor network is updated through the deterministic policy gradient algorithm:
\begin{equation}
\nabla_{\phi} J(\phi)=\mathbb{E}_{s \sim p_{\pi}}\left[\left.\nabla_{a} Q(s, a ; \theta)\right|_{a=\pi(s ; \phi)} \nabla_{\phi} \pi(s ; \phi)\right].
\label{deterministic_pg}
\end{equation}
The motivation here is to update the parameters of the actor network to maximize the critic network output. Therefore, this algorithm renders greedy policies. It is analogous to maximum operator in discrete action spaces. The actor and critic target networks are both updated softly with small $\tau$ values,
\begin{equation}
\theta^{\prime} \leftarrow \tau \theta + (1 - \tau) \theta^{\prime},
\label{soft_update}
\end{equation}
which increases the stability of the agents via slower update. Despite the usage of the target networks, the DDPG algorithm overestimates the Q values especially in large state spaces \cite{lillicrap2019continuous}. 

\subsection{Twin Delayed Deep Deterministic Policy Gradient (TD3)}
One of the successors to DDPG is the TD3 algorithm that introduces regularization on Q-network update to eliminate overestimation \cite{fujimoto2018addressing}. The algorithm utilizes two independently initialized critics and the minimum of which is taken to compute the target for Q-network update. Aside from additional critic, the algorithm also delays the update of the actor to prevent overestimation. The label for the critic update is,
\begin{equation}
    y = r + \gamma \underset{i=1, 2}{\mathrm{min}}Q_{\theta'_{i}}(s', \pi(s'; \phi'_{i})).
\end{equation}

The periodic update of actor and utilization of two critics along with the minimum operator decouple the actor and critic to eliminate the overestimation error. Fujimoto et al. show that TD3  significantly outperforms the DDPG algorithm by correcting the estimation error \cite{fujimoto2018addressing}. Although this method eliminates the detrimental overestimation in the DDPG \cite{lillicrap2019continuous} algorithm, the issue with TD3 algorithm is that the minimum of two critics introduces an underestimation bias \eqref{expected_underestimation}, \cite{DBLP:conf/ictai/HeH20}. 

\subsection{Weighted Delayed Deep Deterministic Policy Gradient (WD3)}

The WD3 algorithm underscores that taking the minimum of two critics leads to underestimation. In response, a weighted average of the minimum and the average value of the two target critic networks is used to update critic networks:
\begin{equation} Q^\prime \leftarrow \beta \min _{i=1,2} Q_{\theta_{i}^{\prime}}\left(s^{\prime}, \tilde{a} \right) \!+\!\frac{1\!-\!\beta}{2} \sum_{i=1}^{2} Q_{\theta_{i}^{\prime}}(s^{\prime}, \tilde{a}), \beta \in (0,1). 
\label{wd3_update}
\end{equation}

They introduce a new hyper-parameter $\beta$ for weighted averaging. While the average of the two critics overestimates, the minimum of the two critics underestimates the true action value function. The motivation is to obtain an accurate estimate of the state action values by combining these extreme points. 

\section{Estimation Bias Phenomenon}

The theoretical study of the estimation bias and the effect of using the minimum operator in continuous action spaces have not been studied extensively. This section focuses on the theoretical and empirical aspects of the estimation bias in continuous state-action spaces. 
\subsection{Overestimation}
\label{subsection:over}

\textbf{Overestimation} indicates the cases when the approximated value exceeds the true value. One reason is the maximum operator in Q-learning \cite{thrun1993issues}. Selecting greedy or near-greedy actions in Q-learning leads to chronic overestimation. Similar to that, the deterministic policy gradient algorithm \eqref{deterministic_pg} causes overestimation in the continuous state-action spaces \cite{fujimoto2018addressing}.

\subsection{Underestimation}
\label{subsection:under}
\textbf{Underestimation} happens when the approximated values are below the true values. In response to the overestimation in the DDPG algorithm, the TD3 algorithm takes the minimum of the estimated values of the two critics \cite{fujimoto2018addressing}. Let $Q^{*}(s, a)$ indicates the true Q value. Assume the estimation errors of the two critics are two correlated Gaussian random variables: 
\begin{equation}
\begin{aligned}
Q_{\theta_{i}}(s, a)-Q^{*}(s, a)=D_{i} \sim \mathcal{N}\left(\mu_{i}, \sigma_{i}^{2}\right), \quad i = 1,2.
\end{aligned}
\end{equation}
Then the expectation of the estimation bias becomes:
\begin{equation}
\begin{aligned}
\mathbb{E}[{\min(D_{1,2})}] \!\!=\!\! 
\mu_{1}\! \Phi\! \left(\frac{\!\mu_{2}\!-\!\mu_{1}}{\sigma}\!\right)\!+\!\mu_{2} \Phi \!\left(\!\frac{\mu_{1}\!-\!\mu_{2}}{\sigma}\!\right)\!-\!\sigma \phi\! \left(\!\frac{\mu_1\!-\!\mu_2}{\sigma}\!\right),  
\label{expected_underestimation}
\end{aligned}
\end{equation}
where $\sigma = \sqrt{\sigma_1^{2} + \sigma_2^{2} - 2\rho\sigma_1\sigma_2}$. The terms $\phi(.), \Phi(.)$ signify the PDF and CDF of the standard normal distribution. He et al. treat the estimation biases of the two critics as independent random variables which is not adequate due to the shared transition data used in the update of the networks \cite{DBLP:conf/ictai/HeH20}. Therefore, we have the term $\rho$ denoting the correlation coefficient. We reveal the underestimation bias in the case of minimizing two estimates \eqref{expected_underestimation}. The next section provides empirical results. 

\subsection{Empirical Demonstration of Estimation Bias}

\begin{figure}[t]
	\centering
	\subfigure{
		\includegraphics[width=1.8in]{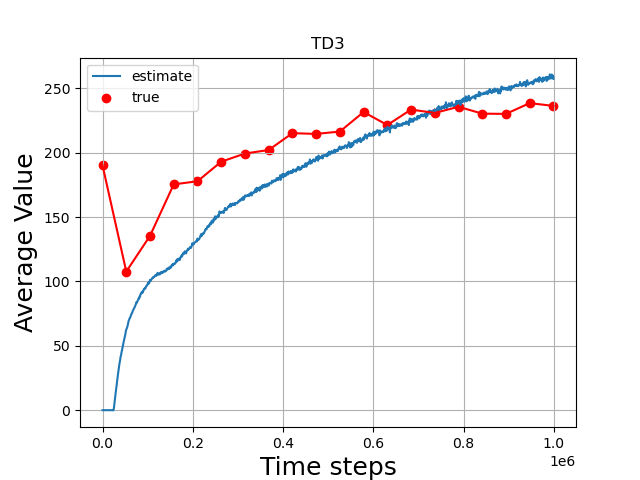}}
	\hspace{-0.30in}
	\subfigure{
		\includegraphics[width=1.8in]{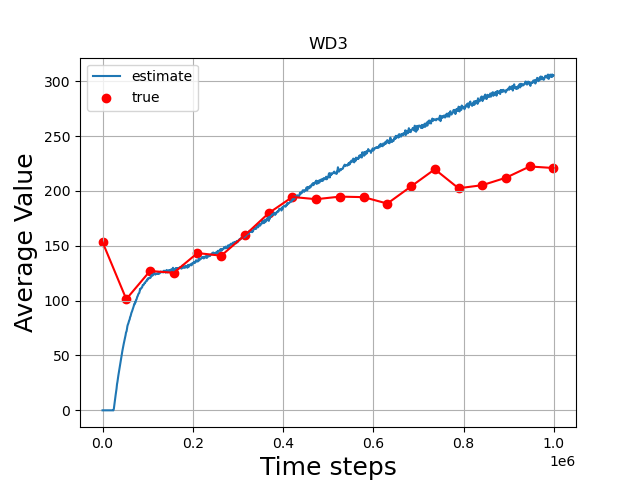}}	
	
	\caption{Measuring the estimation bias in TD3 \& WD3 algorithms in Walker2d-v2 environment averaged over 5 different random seeds} 
	\label{td3_wd3_estimation_bias}
\end{figure}

We observe the estimation bias in the TD3 and WD3 algorithms through the environments in MuJoCo \cite{todorov_erez_tassa_2012}. Fig. \ref{td3_wd3_estimation_bias} shows the state-action value estimations along with the true values for the agents trained on Walker2d-v2 environment. The TD3 agent underestimates at the beginning but it starts to overestimate towards the end. Although the WD3 agent accurately estimates value function at the beginning, it outputs overoptimistic values towards the end. Our simulations verify the theoretical derivations in \ref{subsection:over}, \ref{subsection:under}. We also observe some cases where the agents suffer from overestimation at the beginning of the training process depending on the Xavier initialization networks and the signs of the reward signals\cite{DBLP:conf/ictai/HeH20}. The following section introduces our method.

\section{Adaptive Weighted Delayed Deep Deterministic Policy Gradient (Adaptive WD3)}

To eliminate the estimation bias we propose a new approach which dynamically adjusts the weighting hyper-parameter responsible for balancing between overestimation and underestimation. The following sections provide the key points.

\subsection{Motivation}
Here are some advantages of a dynamic hyper-parameter $\beta$:
\begin{itemize}
\item The hyper-parameter $\beta$ is a tool for balancing the estimation bias by combining two sides. Hence, its value is crucial for the Q-value estimation, therefore also vital for the learning process. Unfortunately, the optimal value for $\beta$ is task-specific, meaning that we may need to have multiple runs for its optimization. In the WD3 paper there are six runs for every environment for different $\beta$ values \cite{DBLP:conf/ictai/HeH20}. In case of a need for more precise value, we must have additional training. Especially in environments where the process of data collection is expensive or slow, tuning $\beta$ becomes unappealing. The need for a hyper-parameter agnostic version of the WD3 algorithm becomes obvious in those cases.  
\item In non-stationary environments where the state transition probabilities are changing, updating $\beta$ may speed up the adaption to the changing true  value function. 
\end{itemize} 

\begin{algorithm}[]
        \caption{Adaptive WD3 algorithm.}
        \begin{algorithmic}
        \STATE Initialize networks $\phi, \theta_{i}, \theta_{i}^{\prime} \leftarrow \theta_{i}, \phi^{\prime} \leftarrow \phi$ for i = 1,2 
        \STATE Initialize $\mathcal{B}, d, \sigma, \tilde{\sigma}, \eta, c, N, T, \beta, \mu, s, t=0$
        \WHILE{$ \textbf{t} < T$}
        \STATE Select $a=\pi(s ; \phi)+\epsilon, \epsilon \sim$ $\mathcal{N}\left(0, \sigma^{2}\right),$ and receive $r,$  $s^{\prime}$
        \STATE Store transition tuple $\left(s, a, r, s^{\prime}\right) \text{to } \mathcal{B}$ 
        \STATE Sample mini-batch of $N$ transitions $\left(s, a, r, s^{\prime}\right)$ from $\mathcal{B}$
        \STATE $\tilde{a} \leftarrow \pi\left(s^{\prime} ; \phi^{\prime}\right)+\epsilon^\prime, \epsilon^\prime \sim \operatorname{clip}\left(\mathcal{N}\left(0, \tilde{\sigma}^{2}\right),-c, c\right)$
        \STATE $y \leftarrow r + \gamma\left(\beta \min _{i=1,2}Q_{\theta_{i}^{\prime}}\left(s^{\prime}, \tilde{a}  \right) + 
        \frac{1-\beta}{2} \sum_{i=1}^{2} Q_{\theta_{i}^{\prime}}\left(s^{\prime}, \tilde{a} \right) \right) $
        \STATE Update critic $ \theta_{i} \leftarrow N^{-1} \sum\left(y-Q_{\theta_{i}}(s, a)\right)^{2}$    
        \IF{$t$ mod $ d $}
            \STATE Update $\phi$ by the deterministic policy gradient:
            \STATE $\nabla_{\phi} J(\phi)=\left.N^{-1} \sum \nabla_{a} Q_{\theta_{1}}\left(s, a \right)\right|_{a=\pi(s ; \phi)} \nabla_{\phi} \pi(s ; \phi) $
            \STATE Update target networks:
            \STATE $\theta_{i}^{\prime} \leftarrow \eta \theta_{i}+(1-\eta) \theta_{i}^{\prime}$, $ \phi^{\prime} \leftarrow \eta \phi+(1-\eta) \phi^{\prime} $
            \ENDIF
            \IF{$s'$ is terminal}
                \STATE Sample last terminal transition $\left(s, a, r, s^{\prime}\right)$
                \STATE $ \tilde{y} \leftarrow \beta \min _{i=1,2}Q_{\theta_{i}'}(s,a) + \frac{1 - \beta}{2} \sum_{i=1}^{2} Q_{\theta_{i}'}(s,a) $  
                \STATE $ \beta \leftarrow \beta - \mu * (r - \tilde{y})$ 
            \ENDIF
        \STATE $ t \leftarrow t + 1 $, $ s \leftarrow s' $
        \ENDWHILE
        \end{algorithmic}
        \label{alg:awd3_alg}
        \end{algorithm}

\subsection{Weighted target update}

As explained in the previous sections, the DDPG and TD3 algorithms suffer from opposite problems named overestimation and underestimation. He et al. proposes the combination of these two effects to achieve a balance through the weighted average of the pair of target critics shown in \eqref{wd3_update}. If $\beta=1$ throughout the training, then the algorithm becomes TD3. Both terms of \eqref{wd3_update} introduce estimation bias. We demonstrate the expectation of the first term with \eqref{expected_underestimation}. The expected value of the second term is $\mathbb{E}\left[D_1 + D_2\right]= 
\mu_{1} +\mu_{2}$ \cite{4403040}. We aim to dynamically adjust the hyper-parameter $\beta$ to zero out the expected value of the estimation bias:
\begin{equation}
\begin{aligned}
\mathbb{E}[D] &= \beta \mathbb{E}[{\min(D_1, D_2)}] + \frac{1 - \beta}{2} \mathbb{E}[D_1 + D_2] = 0.
\label{expected_zero}
\end{aligned}
\end{equation}
Since both critics are trained on the same data and labels, bias values are approximately the same, $\mu_{1} \!\! \approx \!\! \mu_{2}.$ This simplifies \eqref{expected_underestimation}:

\begin{equation}
\begin{aligned}
\beta \left( \frac{\mu_1 + \mu_2}{2} - \frac{
\sigma}{\sqrt{2 \pi}} \right) + (1 - \beta) \frac{\mu_1 + \mu_2}{2} = 0,\\
\beta_{\text{optimal}} = \frac{\sqrt{2 \pi}}{\sigma} \frac{\mu_1 + \mu_2}{2}.
\label{expected_bias}
\end{aligned}
\end{equation} 
Note that the value of $\beta_{\text{optimal}}$ may be more than 1 depending on the values of $\sigma, \mu_1$, and $\mu_2$. The actor network is updated via the gradient of the Q-value estimated by the first critic \eqref{deterministic_pg}. Details are given in Algorithm \ref{alg:awd3_alg}. The next section explains our mechanism to update the weighting hyper-parameter $\beta$. 

\begin{figure*}[t]
	\centering
	\subfigure[Ant-v2]{\includegraphics[width=1.8in,         keepaspectratio]{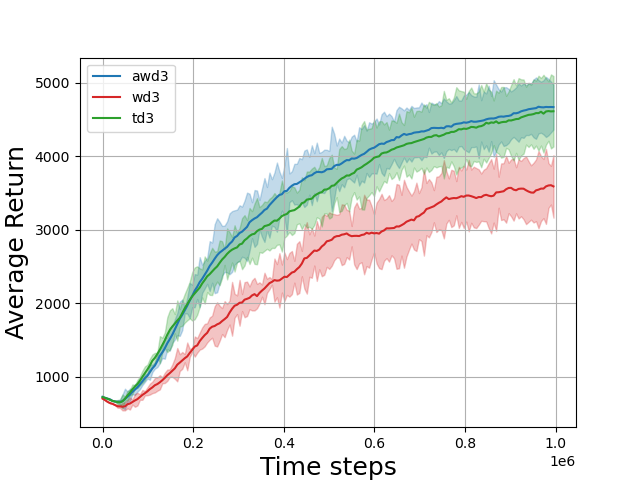}}
	\hspace{-0.30in} 
	\subfigure[Hopper-v2]{
		\includegraphics[width=1.8in, keepaspectratio]{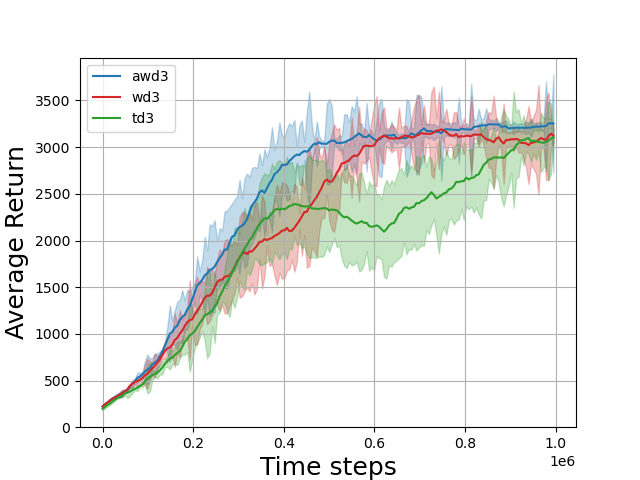}}
	\hspace{-0.30in} 
	\subfigure[Walker2d-v2]{
		\includegraphics[width=1.8in, keepaspectratio]{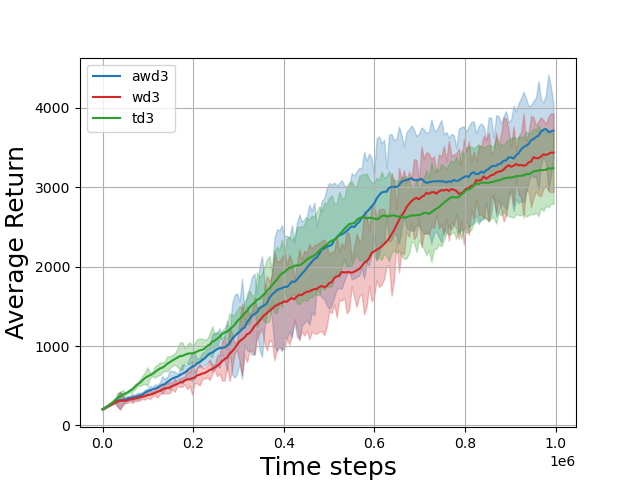}}
	\hspace{-0.30in} 
	\subfigure[BipedalWalker-v3]{
	    \includegraphics[width=1.8in, keepaspectratio]{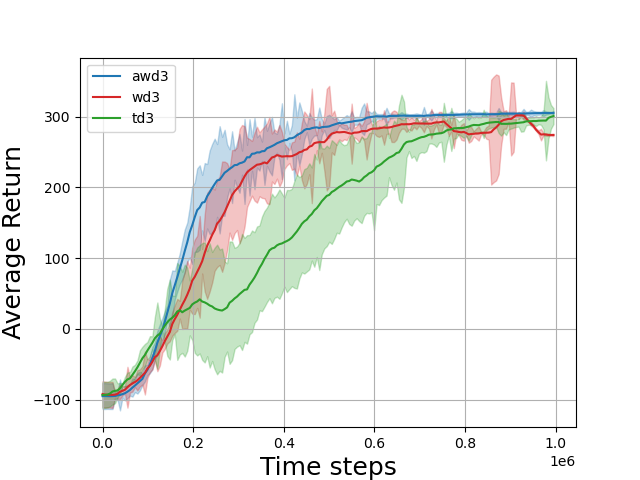}}
	\hspace{-0.35in} 
	\subfigure[Humanoid-v2]{
		\includegraphics[width=1.8in, keepaspectratio]{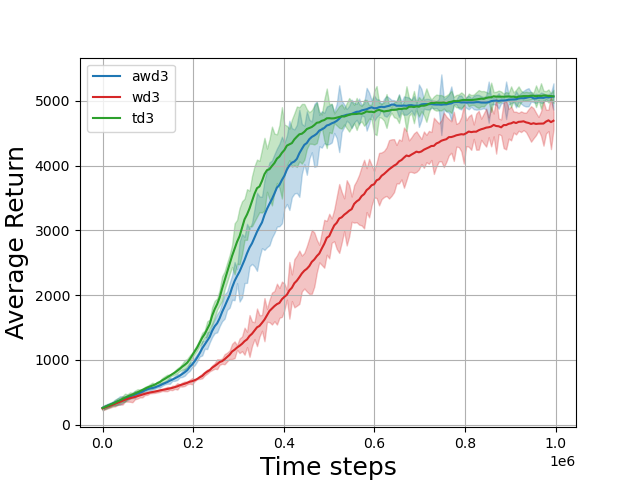}}
	\hspace{-0.30in} 
	\subfigure[InvertedDoublePendulum-v2]{
		\includegraphics[width=1.8in, keepaspectratio]{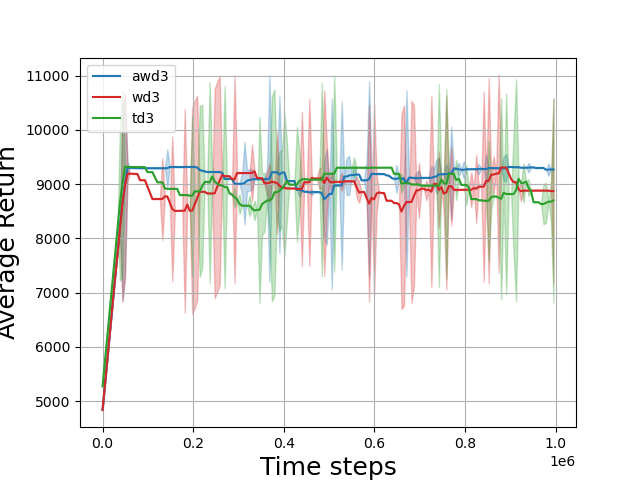}}
	\hspace{-0.30in} 
	\subfigure[LunarLanderContinuous-v2]{
		\includegraphics[width=1.8in, keepaspectratio]{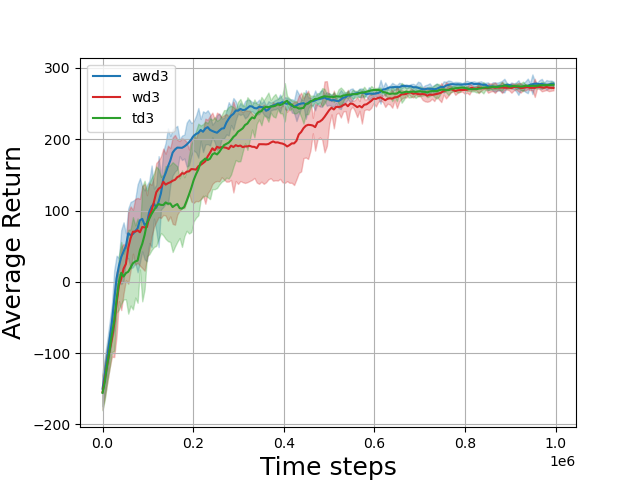}}
	
	\caption{Learning curves for the OpenAI gym continuous control tasks. The shaded region represents half a standard deviation of the
average evaluation over 5 trials. Curves are smoothed.} 
	\label{performance_results}
\end{figure*}

\begin{table*}[t]
\caption{Max Average Return over 5 trials of 1M time-steps. BWalker-v3, Hum-v2, InvDouble-v2, Lunar-v2 stands for BipedalWalker-v3, Humanoid-v2, InvertedDoublePendulum-v2 and LunarLanderContinuous-v2 environments, respectively.}
\begin{center}
\begin{tabular}{c c c c c c c c}
Algs. & Ant-v2 & Hopper-v2 & Walker2d-v2 & BWalker-v3 & Hum-v2 & InvDouble-v2 & Lunar-v2\\ \hline
    TD3 & 4926.3 $\pm$ 859.3 & 3363.6 $\pm$ 154.3 &  3640.3 $\pm$ 738.8 & 308.0 $\pm$ 8.4 & \textbf{5286.0 $\pm$ 66.0} & 9359.7 $\pm$ 0.2 & 288.7 $\pm$ 1.7 \\ \hline
    WD3 & 4157.1 $\pm$ 762.0 & 3328.6 $\pm$ 188.5 &  4217.3 $\pm$ 685.0 & 308.7 $\pm$ 3.0 & 5218.7 $\pm$ 62.7 & \textbf{9359.8 $\pm$ 0.1} & 290.2 $\pm$ 4.3 \\ \hline      
    AWD3  &  \textbf{4948.0 $\pm$ 766.1} & \textbf{3372.6 $\pm$ 114.6} & \textbf{4390.3 $\pm$ 484.9} & \textbf{309.5 $\pm$ 4.2} & 5227.6 $\pm$ 102.4 & \textbf{9359.8 $\pm$ 0.1} & \textbf{293.6 $\pm$ 3.4}  \\ \hline 
\end{tabular}
\end{center}
\end{table*}

\subsection{Updating beta}

Having a dynamic $\beta$ enables us to eliminate the estimation bias without giving precedence over either extreme. The updates of the critic networks are done through the weighted outputs of the target critic networks (\ref{wd3_update}), which we seek to perfectly approximate the true Q function for all state-action pairs in the ideal case. However, we know that the function approximators suffer from the extrapolation error \cite{fujimoto2019offpolicy}. Therefore, in practice, our aim is to estimate Q-values within a negligible error interval for important pairs.

In general, the terminal states play a vital role in training RL agents. Generally, the reward signal given in termination has more magnitude in comparison with other transitions. Moreover, success or failure is sometimes defined on the terminal states. Regarding these, we update the value of $\beta$ to eliminate the estimation bias in the terminal states. 

Recent work shows that the terminations caused by time exceeding  should not be considered as true terminations and be processed carefully \cite{pardo2018time}. Thus, we exclude the terminal states caused by time exceeding in adjusting the value of $\beta$. In addition, we reveal the necessity of expanding the range of $\beta$ both theoretically \eqref{expected_zero} and empirically (Fig. \ref{td3_wd3_estimation_bias}). Therefore, we expand the range of $\beta$. The update rule is as follows: 
\begin{align}
\beta \leftarrow \beta - \mu * (r - \tilde{y}), \quad \beta \in [0, 2.5],
\label{beta_update_rule}
\end{align}
where $r$ is the terminal transition reward, $\tilde{y}$ is the estimated value of the terminal state-action pair and $\mu$ is the learning rate. Since the value of $r$ is unbiased, $\beta$ is guaranteed to converge if the function approximators converge. Next section gives the experimental details.

\section{Experiments}
We evaluate AWD3 and compare its performance with the TD3 and WD3 algorithms via the OpenAI Gym and MuJoCo  environments \cite{todorov_erez_tassa_2012}, \cite{brockman2016openai}. For a fair comparison, we mainly select the common environments mentioned in \cite{fujimoto2018addressing}, \cite{DBLP:conf/ictai/HeH20}. 

\subsection{Implementation Details}

Considering the reproducibility concerns  \cite{henderson2019deep}, we explicitly share our implementation details. We select the hyper-parameters as the same in the TD3 and WD3 algorithms for a fair comparison. We do not manipulate the data coming from the environments and directly input to the networks. We use the default reward setting determined for the environments. We train agents for 1 million time-steps for each setting. For each 5000 time-steps, we stop training and test the agent for 10 episodes. In test mode, agents do not apply exploration noise to the actions taken. In addition to that, transitions experienced in the test mode are not stored in the replay buffer. Dependency to initial parameters is eliminated via randomly sampled actions in the initial 25000 time-steps. In the exploration phase, the networks do not see any updates. We consider the terminations caused by time exceeding normal transitions in updates. This procedure stays the same for all 5 seeds.

The AWD3 algorithm uses the same hyper-parameters across all environments and seeds. Both the actor and two critics have two fully connected feed-forward layers, each layer having 256 neurons. Networks use the ReLU activation function except for the tanh non-linearity in the last layer of the actor network. Adam \cite{kingma2017adam} is the optimizer for the networks. Transitions are uniformly sampled from the replay buffer with a batch size of 100 and the learning rate is 3e-4 for both actor and the two critic networks. Frequency for the actor network and the soft update is $d\!\!=\!\!2$. We set the soft update hyper-parameter $\tau$=5e-3. The critic networks are updated each time step. Exploration noise has a Gaussian distribution $\epsilon \! \sim \! \mathcal{N}(0,0.1)$. After the noise addition, the actions are clipped to be in the action space of the environment. During the update of the critics, a Gaussian noise of $\epsilon\! \sim\! \mathcal{N}(0,0.2)$ is clipped to $[-0.5, 0.5]$ and added to the output of the target actor.

We also introduce new hyper-parameters as a result of the beta value update mechanism. The learning rate for the beta update is 1e-4. Initial beta for environments is taken from \cite{DBLP:conf/ictai/HeH20}. For other environments $\beta$ is initialized as $(\beta\textsubscript{max}+\beta\textsubscript{min})/2$. The update for $\beta$ starts after 100000 time-steps to eliminate the effect of the Xavier initialization and takes place after each episode termination excluding time limit induced termination.  

For a fair comparison, we implement the TD3 and WD3 algorithms through their respective papers without any engineering tricks. Fig. \ref{performance_results} shows the learning curves. These results show that AWD3 outperforms or matches other algorithms and is more robust to catastrophic forgetting.

\begin{figure}[h]
	\centering
	\subfigure[LunarLanderContinuous TD3]{
		\includegraphics[width=1.8in]{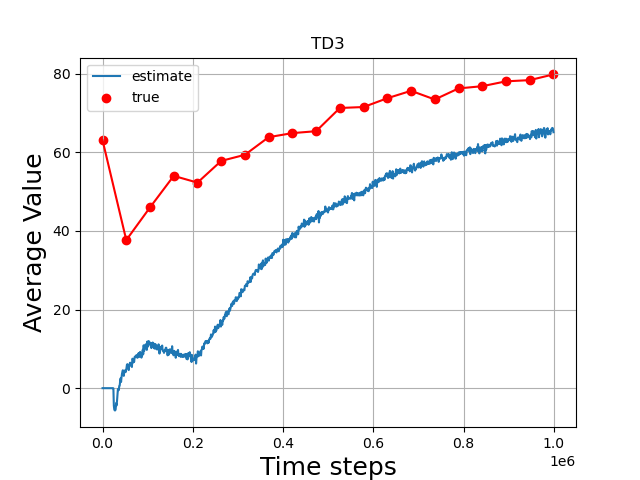}}
	\hspace{-0.35in}
	\subfigure[LunarLanderContinuous WD3]{
		\includegraphics[width=1.8in]{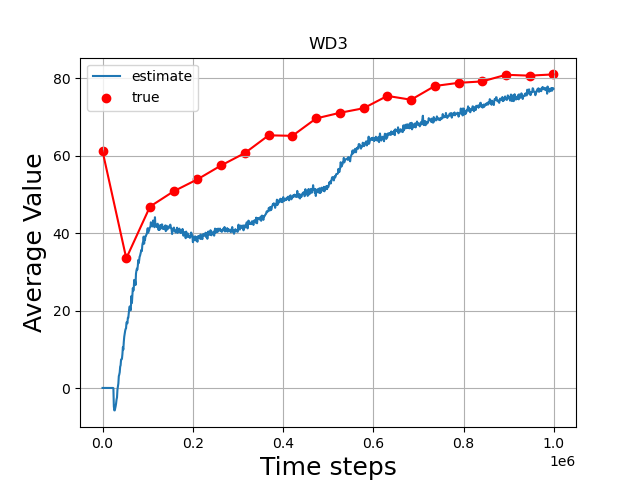}}
	\subfigure[LunarLanderContinuous AWD3]{
		\includegraphics[width=1.8in]{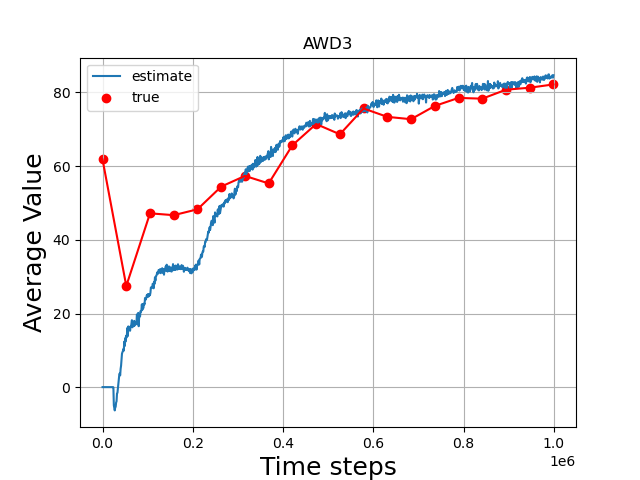}}
	\hspace{-0.35in}
	\subfigure[Walker2d AWD3]{
	\includegraphics[width=1.8in]{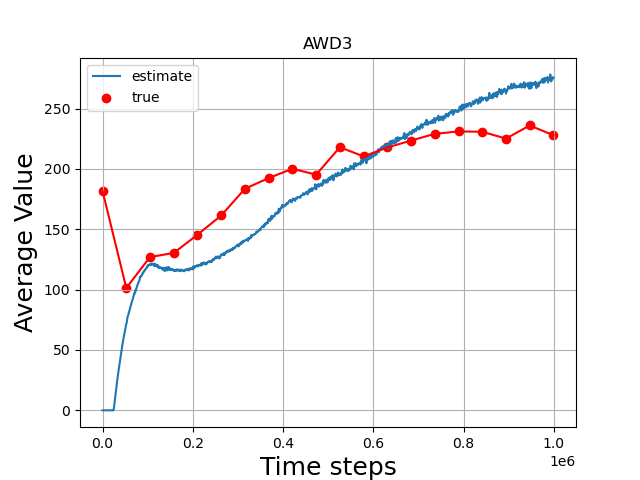}}
	\caption{Empirical demonstration of the estimation bias on continuous control tasks over 5 random seeds. Red line indicates true state-action values whereas blue line is for estimations.} 
	\label{Q_val_estimation}
\end{figure}

\subsection{Q-value estimation}
\label{subsection:q_val}

For a comprehensive comparison, we estimate the Q-values for both three algorithms over 5 seeds. We randomly select 1000 state-action pairs from the replay buffer. For the estimation part, we feed state and action to both critics and calculate estimation according to the respective formulas for each algorithm. For true Q values, we simulate transitions to the terminal states by the Monte Carlo method every 50000 time-step. Fig. \ref{Q_val_estimation} shows the estimated Q-values along with the ground truths. We see that estimating the state-action values by taking the minimum of the two estimators underestimates the ground truth. Our proposal best approximates the Q-values. This advantage increases its performance.

\section{Conclusion}

The estimation bias in continuous action spaces is one of the pivotal setbacks for better performance in value-based RL algorithms. This problem is not completely addressed by the current state-of-the-art deterministic policy gradient algorithms (DDPG, TD3, WD3). Here, we first show the susceptibility of these algorithms to the estimation bias. Then, for active elimination of the estimation bias, we introduce a mechanism to update beta in \eqref{wd3_update}. Through simulations, we verify that our algorithm estimates state action values more precisely better than other algorithms.     

We also show that the range for weighting hyper-parameter $\beta$ in \eqref{wd3_update} may not be sufficient to zero out the expectation of the estimation bias. We verify this derivation with empirical results and show the necessity to expand the range of the weighting hyper-parameter. 

\bibliography{ref}
\bibliographystyle{ieeetr}

\end{document}